\documentclass[conference]{IEEEtran}
\IEEEoverridecommandlockouts
\pdfoutput=1

\usepackage{cite}
\usepackage{amsmath,amssymb,amsfonts}
\usepackage{algorithm}
\usepackage{algorithmic}
\usepackage{graphicx}
\usepackage{svg}
\usepackage{float}
\usepackage{subfigure}
\usepackage{textcomp}
\usepackage{color}
\usepackage{xcolor}
\usepackage{lipsum}
\usepackage{etoolbox}
\usepackage{verbatim}
\usepackage{setspace}
\usepackage{booktabs}

\def\BibTeX{{\rm B\kern-.05em{\sc i\kern-.025em b}\kern-.08em
    T\kern-.1667em\lower.7ex\hbox{E}\kern-.125emX}}

\makeatletter
\patchcmd{\@makecaption}
  {\scshape}
  {}
  {}
  {}
\makeatletter
\patchcmd{\@makecaption}
  {\\}
  {.\ }
  {}
  {}
\makeatother

\title{Staged Depthwise Correlation and Feature Fusion for Siamese Object Tracking\\
}

\author{
\IEEEauthorblockN{Dianbo Ma}
\IEEEauthorblockA{\textit{GSNST} \\
\textit{Kanazawa University} \\
Kanazawa, Japan \\
madb201910@stu.kanazawa-u.ac.jp} \\
\IEEEauthorblockN{Ziyan Gao}
\IEEEauthorblockA{\textit{Information Science} \\
\textit{Japan Advanced Institute of Science and Technology} \\
Nomi, Japan \\
gziyan1237@gmail.com}
\and
\IEEEauthorblockN{Jianqiang Xiao}
\IEEEauthorblockA{\textit{GSNST} \\
\textit{Kanazawa University} \\
Kanazawa, Japan \\
xiaojianqiang0325@hotmail.com} \\
\IEEEauthorblockN{Satoshi Yamane}
\IEEEauthorblockA{\textit{GSNST} \\
\textit{Kanazawa University} \\
Kanazawa, Japan \\
syamane@is.t.kanazawa-u.ac.jp}
}

\begin{document}

\maketitle

\begin{abstract}
In this work, we propose a novel staged depthwise correlation and feature fusion network, named DCFFNet, to further optimize the feature extraction for visual tracking. We build our deep tracker upon a siamese network architecture, which is offline trained from scratch on multiple large-scale datasets in an end-to-end manner. The model contains a core component, that is, depthwise correlation and feature fusion module (correlation-fusion module), which facilitates model to learn a set of optimal weights for a specific object by utilizing ensembles of multi-level features from lower and higher layers and multi-channel semantics on the same layer. We combine the modified ResNet-50 with the proposed correlation-fusion layer to constitute the feature extractor of our model. In training process, we find the training of model become more stable, that benifits from the correlation-fusion module. For comprehensive evaluations of performance, we implement our tracker on the popular benchmarks, including OTB100, VOT2018 and LaSOT. Extensive experiment results demonstrate that our proposed method achieves favorably competitive performance against many leading trackers in terms of accuracy and precision, while satisfying the real-time requirements of applications.
\end{abstract}


\section{Introduction}
Object tracking is one of the fundamental and challenging problems in computer vision community. It has a broad range of applications like autonomous driving, visual surveillance, robotic navigation, and so on. In recent years, deep convolutional neural networks(CNNs) have been drawing vast attention from researchers in related fields. Great progress has been made on object tracking along with the increasing adoption of powerful deep convolutional models. Many successful methods have been proposed for visual tracking \cite{b1,b2,b3,b4,b5,b6,b7}. However, it still suffers from many chanllenges such as heavy occlusions, motion blur, illumination variation, severe deformation, etc. The most popular paradigm for visual tracking has been to learn an embedded template and then compute the similarities \cite{b3,b4}. Specifically, siamese networks are utilized to handle visual tracking through two branches, including one template branch for embedding features extraction and the other search branch for features matching.

Recently, plenty of siamese trackers have been exploited based on SiamFC \cite{b3}. Some approaches focus on modifying the backbones of the models so as to extract more accurate features \cite{b7,b8}. And the other ones manage to tune the loss functions or leverage the detection heads towards receiving the best tracking performance \cite{b5,b6,b9,b10}. The former mainly includes ascending the layers of model, extending the width of network, implementing the extractors of multi-stage structure, using feature fusion and the like. The latter covers employing re-detection/identification for trackers, associating image classification and box regression, optimizing bounding box prediction by mask prediction and miscellaneous ways. 

It is a matter of observation that lower layers of deep models generally produce the edge and corner contours information and higher layers are prone to learn more semantic-specific features \cite{b11}. Low-level features have higher resolution and more position details but less semantics, whereas high-level features have stronger semantic information but lower resolution. Standing on these findings, the hierarchical features from different layers have been effectively integrated in element-wise addition and channel-wise concatenation manners for image classification, object detection, semantic segmentation and other tasks\cite{b12,b13,b14}. In terms of depthwise-style correlation, it is very likely to be confused with a spatially separable convolution. Initially, Sifre \emph{et al.}\cite{b15} develop the depthwise separable convolution (a depthwise convolution followed by a pointwise convolution), which has become an component in popular deep learning frameworks. Since then, a large number of related studies have emerged one after another, such as Network-in-Network\cite{b16}, Inception\cite{b17} and Xception\cite{b18}. These researches demonstrate that both fusing hierarchies of features and using depthwise correlations can generally obtain much better representations and inference results.

\begin{figure*}[tp]
\centering
\includegraphics[scale=0.33]{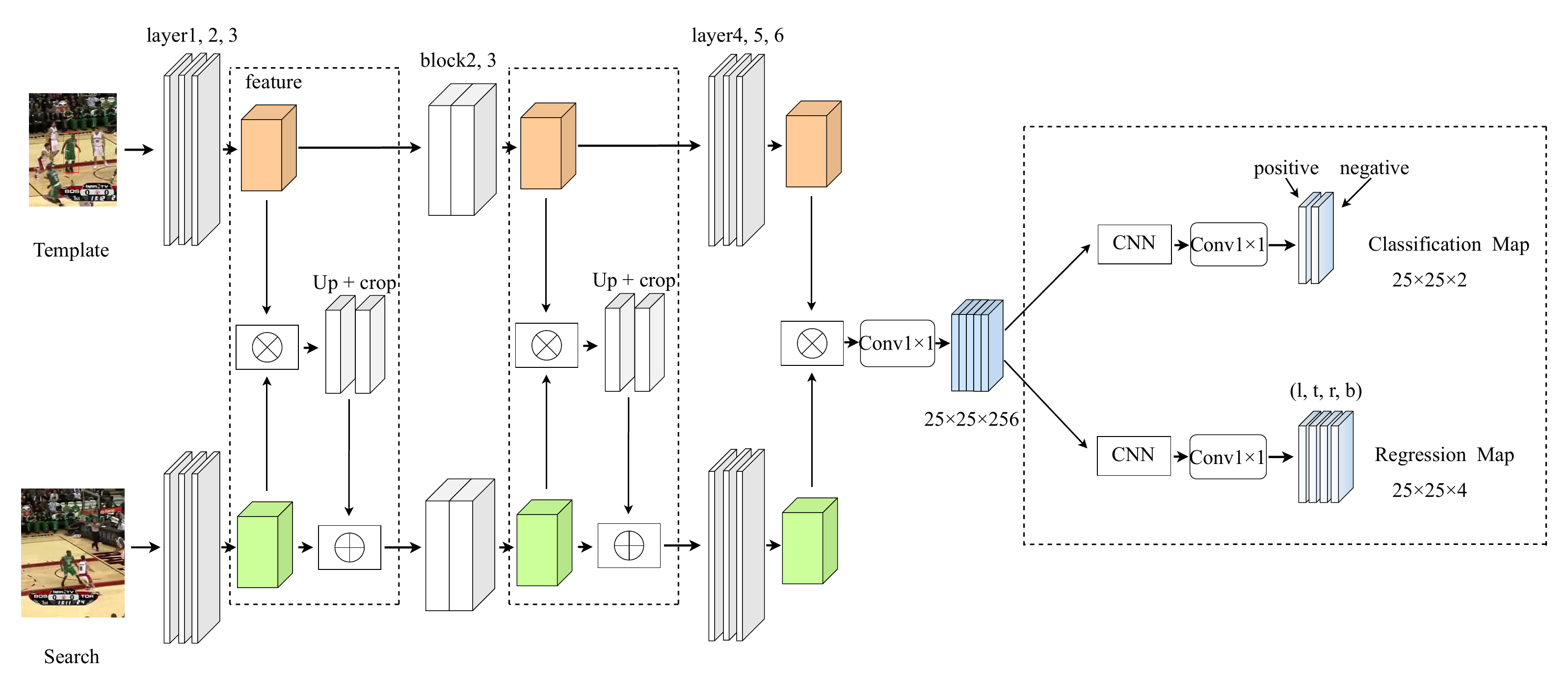}
\caption{An overview of the proposed network. The framework consists of a feature backbone of modified ResNet-50\cite{b13} and a classification-regression prediction head. The correlation-fusion module is integrated into this framework for capturing better appearance representations of the targeted object. The prediction head is adapted from the SiamCAR\cite{b10}. We use the mark $\otimes$ to denote the depthwise correlation, and $\oplus$ for element-wise addition operation.}
\label{Architecture:frame}
\end{figure*}

Motivated by the excellent achievements, especially work relevant to feature fusion, we propose an effective and efficient Siamese framework based \emph{Staged Depthwise Correlation and Feature Fusion Network}, referred to as DCFFNet, to perform more accurate feature learning for visual tracking. The overall architecture is depicted in Fig.\ref{Architecture:frame}. We take the first conv-layer and second, third blocks out of ResNet-50\cite{b13}, and adapt them as part of the backbone of our model. The basic model consists of the first three conventional conv-layers, followed by two conv-blocks (block2 and block3 in ResNet-50), and the last three conventional conv-layers. Then we deposit the depthwise correlation and feature fusion modules (named correlation-fusion) between the third conv-layer and the first conv-block, as well as between the second conv-block and the fourth conv-layer, separately. The correlation-fusion modules are expected to enhance the features and highlight the position of the target. The operation is not only effective for capturing more precise features, but also capable of improving the instability of training and speeding up the convergence of model. When in the training period, we divide the learning processes of the model into a couple of stages. In the beginning, we run 150 epochs on ImageNet VID\cite{b19} dataset to pre-train the parameters for backbone with the Binary Cross Entropy as the loss function following SiamFC\cite{b3}. After that, the optimal parameters are frozen. For better tracking performance, we adopt the classification-regression module from SiamCAR\cite{b10} as the tracking head of our model and refine its parameters on a few datasets, including COCO\cite{b20}, UAV\cite{b21} and the training set of LaSOT\cite{b22}. Extensive experimental results suggest that our model achieves competitive tracking performance, while can strike a trade-off between efficiency and precision. 

Our main contributions can be outlined as follows:

\begin{itemize}
\item We propose a Staged Depthwise Correlation and Feature Fusion Network (DCFFNet) for object tracking. The network contains a core component (i.e., correlation-fusion module), which can extract more accurate features and help the tracker pinpoint the precise location of the target. 

\item As the correlation-fusion modules are embedded in the backbone, the parameters of model are reduced significantly. The training process of the network is much more stable when using stochastic gradient descent (SGD) without tuning the learning rate. Meanwhile, the convergence of model gets faster.

\item The proposed tracker leads to favorable even leading tracking results on major challenging benchmarks while running at a real-time speed, that indicates the potential of the correlation-fusion operation in learning the appearance representations of arbitrary objects. 
\end{itemize}

The following contents of the paper are described in four sections. We review the main related work in Section 2. The proposed DCFFNet is detailed in Section 3. The experiments implementation and results analysis are presented in Section 4. In the end, we draw a conclusion with some discussions in Section 5.

\section{Related Work}
In this section, we revisit the main related work on the object tracking algorithms in the literature and discuss the relations and differences among our proposed method and them.

\textbf{Deep extractor based trackers.} After a successful start in image classification, deep convolutional models have been introduced into visual tracking task, and achieved impressive performances. In \cite{b23}, Ma \emph{et al.} make use of deep learned features in the DCF-based (discriminative correlation filter) model instead of hand-craft features, obtaining substantial increase in performance. Wang \emph{et al.}\cite{b1} propose a fully convolutional network based tracker (FCNT) where a feature-map-selection strategy is adopted to filter distractive and irrelevant feature maps and multi-level features are jointly used with a switch mechanism when tracking object. Nam \emph{et al.}\cite{b2} present a multi-domain network to learn the generic representations of targets from massive video sequences with ground truths for object tracking, and online updating is performed when tracking a target in a new sequence. Fan \emph{et al.}\cite{b24} manage to exploit the recurrent neural networks  (RNNs) to encode the structure of the targeted object, that improves the capabilities of model in both recognizing objects of inter-class and telling similar distractors of intra-class.

\textbf{Siamese trackers.} In the last few years, siamese networks have become a common framework for building many high-quality trackers. Tao \emph{et al.}\cite{b25} design a deep neural network with the siamese structure to learn a matching function from a mass of video sets, and then leverage the knowledgable matching function to search for the targets in candidate patches and return the most similar one. Some of siamese trackers inference the location of targeted object in the subsequent frames by utilizing a learned template to predict a correlation heatmap \cite{b3,b26}. Among them, the most popular algorithm must be SiamFC \cite{b3} since it achieves the state-of-the-art tracking performance at a pretty fast speed with a neat architecture design. Wang \emph{et al.}\cite{b8} present an end-to-end residual attentional siamese networks for object tracking, which not only significantly alleviates the overfitting in the training process, but also improves discriminative ability. The change of aspect ratios is extremely inevitable when tracking object, but the simple response-generation based trackers have not an effective mechanism to handle this case. To solve this issue, Li \emph{et al.}\cite{b5} attach a region proposal network\cite{b27} (RPN) to siamese architecture to constitute an entirely new SiamRPN, showing leading performance. Zhu \emph{et al.}\cite{b28} introduce an effective sampling approach into siamese networks and attempt to offline train the resulting model with more hard negative samples, while a distractor-aware module is designed to transfer the initial embedding to the new video domain for accurate and rubost object tracking.

Feature fusion has been drawn into visual tracking task due to its simplicity and feasibility, and so has depthwise correlation. Among the methods mentioned above, some also use these techniques more or less. The multi-feature fusion module plays a key role in building the Alpha-Refine network\cite{b29}, in which feature fusion and pixelwise correlation\cite{b30} are applied to retain as much spatial information as possible. SPM-Tracker\cite{b31} tackles visual tracking problem as two separate matching stages, namely the CM stage (the coarse matching) and the FM stage (the fine matching). When the two stages in parallel, the matching scores and box regression are fused to reach higher precision of final results. In addition, the FM stage gets the learnt features from conv2 and conv4 layers and aggregates them by concatenation, that is beneficial for promoting the robustness and discriminative power of the network. Guo \emph{et al.}\cite{b10} propose a siamese tracker (SiamCAR), which decouples the visual tracking into two subtasks as classification and regression. In their model, a channelwise concatenation and a depthwise correlation are exerted on the last three residual blocks of the backbone as well as between the template map and the search map, respectively. While so many researches have been exploring feature fusion and depthwise correlation in visual tracking, there still exists a potential of further improvement. We set about exploiting a novel deep siamese model for object tracking. Different from the aforementioned methods, which generally either aggregate the multi-level features and next append a depthwise correlation or launch each kit separately, we operate a depthwise correlation between the template features and search features to generate hidden response maps, and then upsample and crop them to the same size as the search features, finally we fuse resized maps and search maps by an elementwise addition. In this work, we conduct comprehensive experiments on challenging benchmarks, including OTB100\cite{b32}, VOT2018\cite{b33} and LaSOT\cite{b22}. The results demonstrate that our method achieves favorably competitive performance against leading counterparts, whlie running at a real-time speed.

\section{Proposed Method}
To further improve siamese network based trackers, we propose to train a staged depthwise correlation and feature fusion network (DCFFNet) from scratch for visual tracking. Our intuition relys on the insight that depthwise correlation can generate mutiple semantic response maps and feature fusion can develop the strengh of multi-level features. We elaborate the main idea and network architecture below. 

\subsection{Network Architecture}
The proposed DCFFNet framework is based on siamese network as illustrated in Fig.\ref{Architecture:frame}. For an easy understanding of our approach, we first revist the prototype of the basic siamese tracker. In general, siamese trackers entail two weights-shared input branches to extract features from the target patch and the search patch and then treat the template feature maps as kernels to convolve the search feature maps so as to compute similarity reponses. After this pattern, a great variety of advanced methods have been springing up, mainly involving the refinement of feature extractors in frontend and the replacement of prediction heads in backend. The original siamese tracker (e.g., SiamFC\cite{b3}) regards the cropped target patch with ground truth in the fisrt frame as a template $z$, and the local search region in the subsequent frame as an instance $x$. In experiments, pairs of image patches are fed into the parameters-shared CNN extractor and go through the identical transformation $\varphi(\cdot)$ with parameters $\theta$. Finally, the network outputs representative features for two branches. So computing the similarity between template image and instance image is formulated by a naive cross-correlation operation ($\star$) as:
\begin{equation}\label{eq:1}
f_\theta(x,z) = \varphi_\theta(x) \star \varphi_\theta(z) + b
\end{equation}
where $b$ denotes the bias of the similarity score between two images. When $f$ returns a high response value, it implies that the two patches contain the same object. And the candidate with the maximum score locates the targeted object in current frame.

In SiamFC\cite{b3}, the tracking framework takes in the shallow yet powerful convolutional neural networks, i.e., AlexNet\cite{b34}, as the backbone to learn appearace representations of the object, achieving an amazing tracking trade-off between accuracy and speed. It has been proven that much deeper networks can significantly boost the inference capabilities of the siamese-based tracking methods\cite{b6,b7}. However, directly replacing the shallow CNNs in the frontend with deeper ones barely improves the performance of siamese tracker even worse. According to proir observations, we rebuild the feature extractor of siamese tracking model based on ResNet-50\cite{b13}. We choose only the first convolutional layer and second, third convolutional blocks from ResNet-50 and adapt these components to our proposed network. The overall network architecture consists of the first three conv-layers, intermediate two conv-blocks, the last three conv-layers and a tracker head. The correlation fusion module that is the core component serves as a retainer and is inserted into network framework for learning more discriminative representations.

\subsection{Correlation Fusion Module}
The common cross-correlation is an indispensable operation for the original siamese-based trackers to compute the consistencies between template images and instance images. SiamFC\cite{b3} sets up a cross-correlation layer on top of fully convolutional networks to produce response score for measuring the gap between target object and candidate region. In SiamRPN\cite{b5}, the extracted features from CNNs are split into two branches, i.e., classification branch and regression branch, and a pair-wise correlation is imposed on both two branches. But this approach increases the channels of template branch in RPN\cite{b27} exaggeratively, that is bound to cause the imbalance of parameter distribution over the feature extractor and the RPN subnetwork. Thus, SiamRPN++\cite{b6} introduces the lightweight depthwise correlation (DW-XCorr) to embed associated imformation between template and instance features and meanwhile to reduce parameters in correlation layer. Each channel of the generated multi-channel responses maintains the same semantic information. We clone the depthwise correlation used in SiamRPN++\cite{b6} to the middle layers of our siamese networks. Different from previous methods, we execute depthwise correlation between bilateral latent outputs in a staged manner. The third conv-layer and second conv-block output the feature maps of size 87$\times$87 and 44$\times$44 in search branch, respectively. The resulting multiple semantic maps are reshaped into the same size as current feature maps in search branch by a trilinear interpolation operation so that the following feature aggregation can be fulfilled smoothly. 

Aggregating multi-level features is supposed to develop the advantages of different convolutional layers. The emergence of deeper convolutional neural networks makes feature association of different layers much easier. In particular, ResNet\cite{b13} introduces a  residual learning approach, which enables neural networks to achieve the record-breaking performance with a depth of up to 152 layers. This approach sovles the problem that deep networks are different to train. ResNet\cite{b13} is a representative work that makes full use of multi-level features. As expected, it has been widely used in deep model designing for various visual tasks. Following the prior methods, we adjust the portions of ResNet-50\cite{b13} to our framework. In addtion, we exert feature aggregation outside residual blocks. 

In our scheme, the depthwise correlation and feature fusion are integrated into a flexible module. The operation is carried out twice throughout the whole architecture. For more detailed settings see Fig.\ref{Architecture:frame}.

\subsection{Objective Function}
In pre-training process, we adopt the same loss function as in SiamFC\cite{b3}. We train the proposed model offline for 150 epochs with randomly sampled image pairs ($x$,$z$) and corresponding ground-truth $\mathbb{Y}$ on ImageNet\cite{b19} datasets. In \eqref{eq:1}, the higher response value $f$ returns, the more likely two patches are to depict the same object. The goal is naturally transformed into minimizing the distance between $f$ and label $\mathbb{Y}$. Therefore, the logistic loss $l$ can be expressed as: 
\begin{equation}\label{eq:5}
\arg \min_{\theta} \sum_{(x,z,\mathbb{Y})} l(f_\theta(x,z),\mathbb{Y}) 
\end{equation}
where $\theta$ denotes the parameters of the convolutional networks.

\begin{figure*}[tp]
  \centering
  \includegraphics[width=9cm]{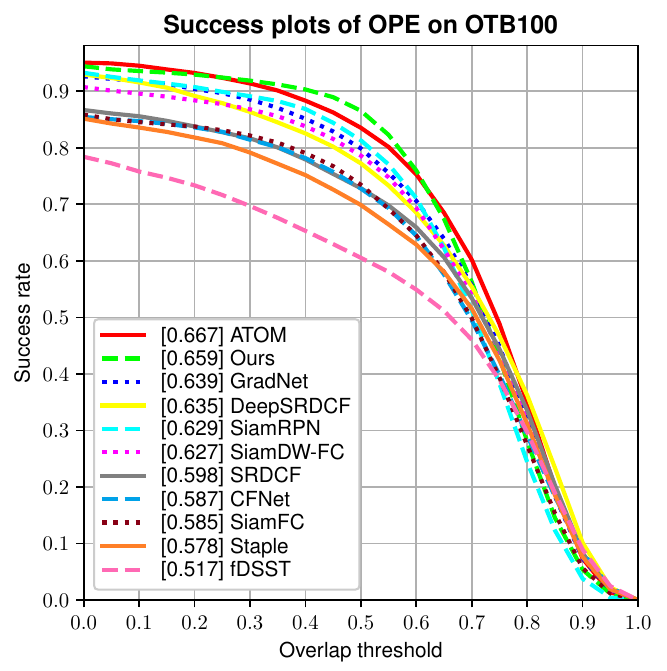}
  \includegraphics[width=9cm]{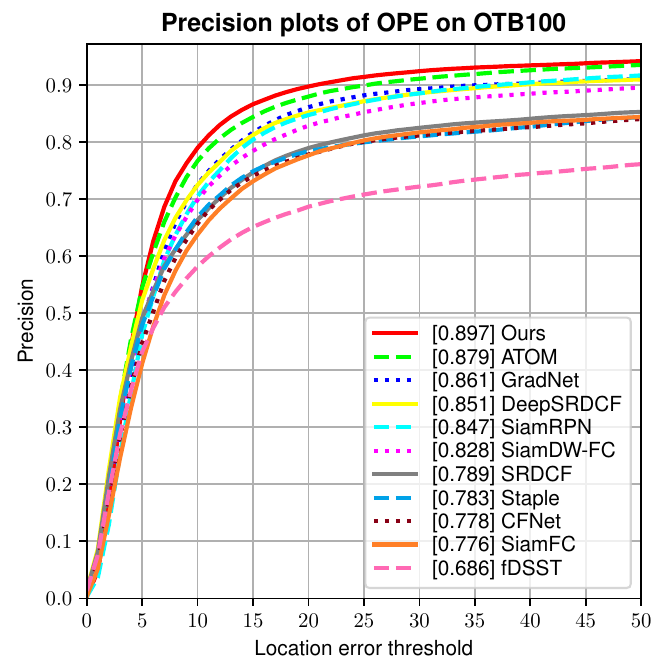}
  
  \caption{Success and precision plots on OTB100\cite{b32} benchmark. The legend showes the area-under-curve (AUC) score and the distance precision score at the threshold of 20 pixels for each tracker. Better viewed in color.}
  \label{OTB:results1}
\end{figure*}

After the initial training, we select the optimal parameter configurations. To achieve the best inference capability, the prediction head presented in SiamCAR\cite{b10} is combined with the feature extractor in DCFFNet for the next refining process. However, the output size from the convolutional backbone is not compatible with the input size of external prediction head. As a result, we fix the convolutional networks except for the last two layers, and perform the depthwise correlation on the last output maps of two branches to generate suitable size. We fine-tune the parameters of the last two layers in conjunction with the prediction head on well-known datasets including COCO\cite{b20}, UAV\cite{b21} and the training set of LaSOT\cite{b22}.

The prediction head addresses the object tracking through two branches: a classficator to discriminate the category for every location, and a regressor to predict the bounding box for this location. Given a response map $M_{w \times h \times c}$ from the preceding extractor, the classficator generates a classification map $C_{w \times h \times 2}$ and the regressor generates a regression map $R_{w \times h \times 4}$. The width, height and channels of the output feature maps are denoted as $w$, $h$ and $c$ respectively. Every location $(x,y)$ on the feature map $C_{w \times h \times 2}$ is recognized as a positive sample when its corresponding location $(x^*,y^*)$ on the search image is included in the ground truth box, a negative sample if not. For each positive sample $(x,y)$ on the feature map $R_{w \times h \times 4}$, the regressor computes the distances from its corresponding location $(x^*,y^*)$ to the four sides of ground truth box on the search image. Here we denote the distances as a 4D vector $\mathbf{\hat{t}} = (l^*,t^*,r^*,b^*)$. Therefore, the regression targets of location $(x,y)$ can be written as:
\begin{equation}\label{eq:6}
\begin{aligned}
l^* = x^* - x_0, ~~~ t^* = y^* - y_0, \\
r^* = x_1 - x^*, ~~~ b^* = y_1 - y^*
\end{aligned}
\end{equation}
where $(x_0,y_0)$ and $(x_1,y_1)$ indicate the top-left and bottom-right coordinates of ground-truth annotations in the original search image, respectively.

We find that it is not necessary for our networks to use ceterness loss, and accordingly we remove the ceterness branch used in SiamCAR\cite{b10}. We just leverage the straightforward Cross-Entropy Loss for classification task and the IOU Loss\cite{b35} for regression task. To some extent, the modification helps to reduce the complexity of the networks and make training process more stable. So we can summarize the regression loss as:
\begin{equation}\label{eq:7}
  \begin{split}                
  L_{reg} &= \frac {1}{N_{pos}} \sum_{x,y} \mathbf{1}_{\{c_{x,y}^* > 0\}} L_{IoU} (\mathbf{t}_{x,y},\mathbf{\hat{t}}_{x,y})  \\    
\end{split}
\end{equation}
where $\mathbf{1}_{\{\cdot\}}$ is an indicator function that gets 1 when the condition in brackets is satisfied, gets 0 if not, and $L_{IoU}$ denotes the IoU loss \cite{b35}. And $c_{x,y}^*$ takes 1 when location $(x,y)$ is recognized as a positive sample, otherwise takes 0. 

Finally, we simplify the overall objective function as:
\begin{equation}\label{eq:8}
L_{total} = L_{cls} + L_{reg} 
\end{equation}
where $L_{cls}$ is the common binary cross-entropy loss for classification result. The loss is enabled in the fine-tuning training.

\section{Experiments}
In this section, we evaluate our model and compare it with the state-of-the-art trackers on the major benchmarks including OTB100\cite{b32}, VOT2018\cite{b33} and LaSOT\cite{b22}. We present extensive experiment results and analysis to verify the effectiveness of the proposed algorithm. In addtion, we carry out ablation studies to show the inference capabilities of different setting options. Our tracker achieves the leading results and run at a speed of 39 fps, which meets the real-time requirements of applications.

\begin{table*}[tp]
  \caption{Detailed comparison results on VOT2018\cite{b33} dataset. In the table, we compare our tracker with nine state-of-the-art trackers in terms of expected average overlap (EAO), accuracy score (A) and robustness (R). The best two results are presented in \textcolor{red}{red} and \textcolor{blue}{blue} fonts, respectively.}
  \label{tab1}
  \begin{center}
  
  \begin{tabular}{l ccccccccc r}
    \midrule\noalign{\smallskip}
    & \textbf{UPDT} & \textbf{LADCF} & \textbf{RCO} & \textbf{SiamFC} & \textbf{SPM} & \textbf{SiamRPN} & \textbf{SiamFC++} & \textbf{ATOM} & \textbf{SiamRPN++} & \textbf{Ours}  \\ 
    \noalign{\smallskip}\hline\noalign{\smallskip} 
    \textbf{EAO} & 0.379 & 0.389 & 0.376 & 0.188 & 0.338 & 0.384 & 0.400 & 0.401 & \textcolor{blue}{0.414} & \textcolor{red}{0.431}  \\
    \textbf{A} & 0.536 & 0.503 & 0.507 & 0.503 & 0.580 & 0.588 & 0.556 & \textcolor{blue}{0.590} & \textcolor{red}{0.600} & 0.586  \\
    \textbf{R} & 0.184 & \textcolor{blue}{0.159} & \textcolor{red}{0.155} & 0.585 & 0.300 & 0.276 & 0.183 & 0.204 & 0.234 & 0.176  \\
    \noalign{\smallskip}\midrule
  \end{tabular}

  \end{center}
\end{table*}

\begin{table*}[tp]
  \caption{Performance comparison with the baseline and state-of-the-art algorithms on the LaSOT\cite{b22} benchmark in both success and fps measures. The numbers in \textcolor{red}{red} and \textcolor{blue}{blue} denote the best and second best results under the AUC metric, respectively.}
  \label{tab2}
  \begin{center}
  
  \begin{tabular}{l cccccccccc r}
    \midrule\noalign{\smallskip}
    & \textbf{CSRDCF} & \textbf{CFNet} & \textbf{Staple} & \textbf{ECO-hc} & \textbf{ECO} & \textbf{SiamFC} & \textbf{VITAL} & \textbf{MDNet} & \textbf{SiamRPN} & \textbf{SPM} & \textbf{Ours}  \\ 
    \noalign{\smallskip}\hline\noalign{\smallskip} 
    \textbf{AUC} & 0.244 & 0.275 & 0.243 & 0.304 & 0.324 & 0.336 & 0.390 & 0.397 & \textcolor{blue}{0.432} & \textcolor{red}{0.471} & 0.427  \\
    \textbf{FPS} & 13 & 75 & 80 & 60 & 8 & 86 & 1 & 1 & 160 & 120 & 39  \\
    \noalign{\smallskip}\midrule
  \end{tabular}

  \end{center}
\end{table*}

\subsection{Training Details}
The backbone of proposed DCFFNet is designed based on the ResNet-50\cite{b13}. And the prediction heads are adapted from SiamFC\cite{b3} and SiamCAR\cite{b10}. We pre-train our framework with the first prediction head from scratch on ImageNet VID\cite{b19} (4417 videos), which is a very appropriate initialization way to various tasks. Afterwards the parameters of the backbone are frozen except for the last two layers. For fine-tuning, we train the network with the second prediction head on multiple datasets: COCO\cite{b20}, UAV\cite{b21} and LaSOT\cite{b22}. We follow the official evaluation protocol and only take the training splits (1120 videos) of LaSOT. Except where specified otherwise, each of four datasets is split into training and validation sets in a $90\%-10\%$ ratio (separately). Considering enough data for the second stage, we utilize grayscale, translate, flip, and scale augmentations during only initialization.

The proposed tracker obtains the inputs of single scale by cropping and scaling the image pairs randomly sampled from abundant video sequences to 127 pixels for the template branch and 255 pixels for the search branch. We train our tracker for total 450 epochs with the stochastic gradient descent (SGD) and a minibatch of 72 (32 pairs per GPU), of which the first 150 epochs are conducted for initialization with a learning late of 0.01. When fine-tuning the model, we run 150 epochs, 100 epochs and 50 epochs with learning rates of 0.01, 0.001 and 0.0001 respectively, and the weight decay is set to 0.0005 and the momentum is set to 0.9 all the way. For the whole training process, all experiments are performed on a computer with 128G memory, a 4.60GHz i9-10980XE CPU and two Nvidia RTX 3090 GPUs.

\subsection{Quantitative Evaluation}
\textbf{OTB100 Dataset.} The dataset consists of 100 video sequences with 11 different attributes. It has been serving as the most popular benchmark since its release and made a significant difference in the development of visual tracking algorithms. On the benchmark, we obey the conventional evaluation metrics, that is, the one-pass evaluation (OPE), to compare our method and other trackers. The OPE metrics measure the performances of algorithms on the distance precision and the area-under-curve (AUC) of overlap success plots.

We execute the contrast experiments between our tracker and the recent state-of-the-art trackers, including SiamFC\cite{b3}, CFNet\cite{b4}, SiamRPN\cite{b5}, SiamDW-FC\cite{b7}, Staple\cite{b26}, GradNet\cite{b36}, ATOM\cite{b37}, SRDCF\cite{b38}, DeepSRDCF\cite{b39} and fDSST\cite{b40}. Fig.\ref{OTB:results1} shows the experiment results of diferent methods on the dataset. On the whole, our tracker achieves favorably competitive results compared to ten trackers and exceeds the most of these methods. In particular, our method achieves the best performance on the precision plot. Specifically, our tracker outperforms SiamFC\cite{b3} and SiamRPN\cite{b5} by $7.4\%$ and $3\%$ gains on the success rate plot respectively, certifying the effectiveness of the proposed DCFFNet for learning better feature representations.

\begin{table}[tp]
  \caption{Ablation analysis of our tracker with different setting options on the OTB100\cite{b32} dataset.}
  \label{tab3}
  \begin{center}
  
  \begin{tabular}{ccccc}
    \bottomrule\noalign{\smallskip}
    & Baseline & CF-1st & CF-2nd & CF-double \\ 
    \noalign{\smallskip}\hline\noalign{\smallskip}
    AUC & 0.612 & 0.633 & 0.621 & 0.659  \\ 
    
    Precision & 0.793 & 0.861 & 0.850 & 0.897 \\
    \noalign{\smallskip}\hline\noalign{\smallskip}
    FPS & 106 & 57 & 79 & 39 \\ 
    \noalign{\smallskip}\bottomrule

  \end{tabular}
  \end{center}
\end{table}

\textbf{VOT2018 Dataset.}
We evaluate our algorithm on the Visual Object Tracking chanllege VOT2018\cite{b33} benchmark, which contains 60 videos with diverse challenging scenarios. Compared to other tracking datasets, VOT2018 can render a more high-precision localization evaluation by applying the rotatable rectangles to label the ground truth box. On the dataset, the expected average overlap (EAO) is employed to measure the overall performance of trackers in terms of accuracy and robustness. 

We abide the evaluation guideline, specified in the protocol, to repeat each experiment over 15 times and compute the average. When the trackers fail to detect and lose the target, they are reset by the evaluation mechanism. Table \ref{tab1} reports the comparison results of our tracker with nine top-ranked trackers, including SiamFC\cite{b3}, SiamRPN\cite{b5}, SiamRPN++\cite{b6}, SiamFC++\cite{b9}, SPM\cite{b31}, RCO\cite{b33}, ATOM\cite{b37}, UPDT\cite{b41} and LADCF\cite{b42}. Among these algorithms, our approach reaches the best EAO score and is slightly lower than the best two trackers in accuracy score. This says our tracker is capable of recognizing the targeted object accurately and is also rubust to variable and challenging cases during the tracking process.

\textbf{LaSOT Dataset.}
We also evaluate our model on the Large-scale Single Object Tracking (LaSOT \cite{b22}) benchmark, which gathers 1400 video sequences with up to 3.52 million frames (an average video length of over 2500 frames). It is by far the largest dataset for single object tracking. The dataset adopts the one-pass evaluation to measure the performance of trackers on two criteria (success and precision). Because the training split of LaSOT has been employed to fine-tune the proposed network, we follow the evaluation protocol to use only the testing split which contains 280 sequences. 

On LaSOT, we consider ten top-performing trackers, including MDNet\cite{b2}, SiamFC\cite{b3}, CFNet\cite{b4}, SiamRPN\cite{b5}, Staple\cite{b26}, SPM\cite{b31}, CSRDCF\cite{b43}, ECO/ECO-hc\cite{b44} and VITAL\cite{b45}. For fair comparision, different families of tracking methods get involved in the experiments. The comparison results is stated in Table \ref{tab2}. Our tracker obtains favorably competitive performance against the recent prominent trackers on the benchmark. In AUC measure, though our method is inferior to the first and second best trackers by a small margin, it is much better than the remaining trackers. Compared to SiamFC\cite{b3}, also trained on ImageNet VID\cite{b19} for initialization, our DCFFNet accomplishes a remarkable improvement in accuracy, while still running at over 30 FPS.

\subsection{Ablation Studies}
For an in-depth probe, we do the ablation studies to confirm the effectiveness and flexibility of the correlation-fusion module (short for CF module). We implement the comparision experiments on OTB100\cite{b32}. The basic symbols and meanings are listed below. $(1)$ `Baseline' means the backbone of tracker is the single modified ResNet-50\cite{b13} with no correlation-fusion module. $(2)$ `CF-1st' says the DCFFNet has only one correlation fusion module, inserted between the third layer and first block of backbone. $(3)$ `CF-2nd' indicates that the DCFFNet is equipped with only one correlation fusion module, placed between the second block and fourth layer of backbone. $(4)$ `CF-double', just as the name suggests, is the combination of the CF-1st and CF-2nd. Table \ref{tab3} presents the comparison results. Both CF-1st and CF-2nd boost the performance of the network over the baseline, of which the CF-1st is better than the CF-2nd in terms of accuracy and precision. The CF-double obtains the best results compared with other three options. These results indicate the CF module is capable of improving the learning abilities of neural networks and better at extracting the hidden pattern in lower layers of neural networks.

\section{Conclusions}
In this work, we propose the Staged depthwise correlation and Feature Fusion Network (DCFFNet) for visual tracking. We exploit the potential of correlation-fusion method in learning more exact appearace representations of arbitrary objects. The proposed network is pre-trained on ImageNet VID from scratch and refined on COCO, UAV and the training split of LaSOT. DCFFNet is qualified to capture the precise location of targeted object by combining depthwise correlation and feature fusion. In comprehensive experiments, our algorithm achieves favorably competitive results against the recent leading trackers on OTB100, VOT2018 and LaSOT, while running at a real-time speed of 39 fps.

\end{document}